\documentclass[runningheads]{llncs}
\usepackage{graphicx}
\usepackage{xcolor}
\usepackage{booktabs}%
\usepackage{multicol}
\usepackage{graphicx}
\usepackage{tcolorbox}

\usepackage{etoolbox} 
\makeatletter
\patchcmd{\ps@headings}{\rlap{\thepage}}{}{}{}
\patchcmd{\ps@headings}{\llap{\thepage}}{}{}{}
\makeatother
\begin{document}
%
\title{Large Language Models for Sentiment Analysis to Detect Social Challenges: A Use Case with South African Languages}
%
\titlerunning{Large Language Models for Sentiment Analysis to Detect Social Challenges}
%

\author{Koena Ronny Mabokela\inst{1}\orcidID{0000-0002-8058-969X}, Tim Schlippe\inst{2}\orcidID{0000-0002-9462-8610}, Matthias Wölfel\inst{3}\orcidID{0000-0003-1601-5146}}
\authorrunning{K. R. Mabokela, T. Schlippe and M. Wölfel}
\institute{University of Johannesburg, South Africa
\and
IU International University of Applied Sciences, Germany
 \and
Karlsruhe University of Applied Sciences, Germany\\
\email{krmabokela@gmail.com}}
\maketitle   

\begin{abstract}
Sentiment analysis can aid in understanding people's opinions and emotions on social issues. In multilingual communities sentiment analysis systems can be used to quickly identify social challenges in social media posts, enabling government departments to detect and address these issues more precisely and effectively. Recently, large-language models (LLMs) have become available to the wide public and initial analyses have shown that they exhibit magnificent \textit{zero-shot} sentiment analysis abilities in English. However, there is no work that has investigated to leverage LLMs for sentiment analysis on social media posts in South African languages and detect social challenges. Consequently, in this work, we analyse the \textit{zero-shot} performance of the state-of-the-art LLMs GPT-3.5, GPT-4, LlaMa~2, PaLM~2, and Dolly~2 to investigate the sentiment polarities of the 10~most emerging topics in English, Sepedi and Setswana social media posts that fall within the jurisdictional areas of 10~South African government departments. Our results demonstrate that there are big differences between the various LLMs, topics, and languages. In addition, we show that a fusion of the outcomes of different LLMs provides large gains in sentiment classification performance with sentiment classification errors below 1\%. Consequently, it is now feasible to provide systems that generate reliable information about sentiment analysis to detect social challenges and draw conclusions about possible needs for actions on specific topics and within different language groups.

\keywords{AI for Social Good  \and Sentiment Analysis \and Natural Language Processing \and South Africa \and Large-Language Models \and LLMs.}
\end{abstract}
\section{Introduction}

Artificial intelligence (AI) has revolutionised different areas and is now also addressing societal issues~\cite{Tomaev2020AIFS}, with a focus on achieving the United Nations’ Sustainable Development Goals (SDGs)~\cite{SDGs:2022}. In South Africa, the National Development Plan aligns 74\% with the SDGs, emphasizing job creation, poverty elimination, inequality reduction, and inclusive economic growth~\cite{CountryReportSouthAfrica}, with various government departments mandated to support these goals. 

Sentiment analysis involves automatically detecting and classifying sentiments from textual data into categories like \textit{negative}, \textit{neutral}, or \textit{positive}~\cite{Wankhade:2022} with the help of AI and natural language processing. Applying sentiment analysis to online texts posted by citizen of a specific population can help to automatically and rapidly find and tackle social challenges in this population~\cite{Kiritchenko2018ExaminingGA}.

While sentiment analysis tools are widely available for English, which is spoken by only 19\% of the global population, it is crucial to extend these tools to other languages, particularly in multilingual societies like South Africa~\cite{Statista:2022}. With 12~official languages, including low-resource Niger-Congo Bantu languages~\cite{mabokela-etal-2023-investigating,SACensus2022}, there is a need for sentiment analysis applications that can process texts in these languages to effectively detect and address social challenges.

However, for most African languages it is very challenging to build sentiment analysis systems due to the limited availability of natural language processing corpora. Furthermore, only experts can deal with the complex algorithms required for training and fine-tuning traditional sentiment analysis systems. Given that state-of-the-art LLMs have the potential to address these problems through their growing capabilities and ease of use through prompting, particularly in \textit{zero-shot}, we investigated their performance for sentiment analysis in African languages. Consequently, we automatically analysed the following government departments related topics with the help of state-of-the-art LLMs: \textit{employment}, \textit{sanitation}, \textit{police service}, \textit{education}, \textit{health}, \textit{small business}, \textit{transport}, \textit{home affair}, \textit{rural development}, and \textit{agriculture}. For each topic, we had the LLMs classify corresponding social media posts into the 3~categories \textit{negative}, \textit{neutral} and \textit{positive} and compare the LLM performances to sentiment analysis systems. 

For our study, we collected 16,000~social media posts from  X\footnote{former Twitter}, i.e.\ tweets, in the three languages English, Sepedi and Setswana containing our government department related topics: the \textit{SAGovTopicTweets} corpus. 

Our contributions are:
\begin{itemize}
    \item We analyse state-of-the-art LLMs' sentiment analysis performances to detect social challenges in 3 South African languages.
    \item We leveraged the \textit{SAGovTopicTweets} corpus to evaluate the performances of the tested LLMs, covering 10 South African government departments-related topics.
    \item Our results can be used as recommendations for the South African government departments to improve the social challenges identified on social media.
\end{itemize}

In the next section, we will describe related work. The experimental setup of our collection and sentiment analysis of tweets in English, Sepedi and Setswana will be presented in Section~3. In Section~4, we will demonstrate the results of our experiments. Finally, we will summarise our work and indicate possible future steps.

\section{Related Work}

AI for Social Good is a growing field of study that deals with the development of AI-based methods to enhance community well-being~\cite{Musikanski2020ArtificialIA}. \cite{AIU4SG:2020} provide a comprehensive analysis of various approaches, use cases, and examples within this field. Many AI for Social Good applications employ learning, reasoning, heuristic search, and problem-solving algorithms~\cite{AIU4SG:2020}, which are widely utilised by numerous organizations and economic sectors~\cite{Corneliu2022}. There is a significant demand for AI applications that benefit society, as they have the potential to address numerous challenges~\cite{Hager2019ArtificialIF}.

In the field of NLP for social good,~\cite{Kiritchenko2018ExaminingGA} utilised sentiment analysis to automatically detect gender and race bias.~\cite{Kaur:2020} explored sentiment analysis techniques to classify five main social issues: corruption, violence against women, poverty, child abuse, and illiteracy, collecting English tweets and applying machine learning algorithms. 
\cite{Makuwe:2023} investigated sentiment analysis for the African language Shona and reports the Shona speakers' sentiments on different topics. 
Text data from microblogging platforms like X (former Twitter) is often used due to its situational information, topic diversity, and range of sentiments, e.g.\ by~\cite{Go2009,Indriani2020}. Various studies have examined methods for collecting tweets~\cite{Go2009,paktwitter2010,Agarwal2012EndtoEndSA,nakov2016semeval,Indriani2020}.  

For automatic sentiment analysis, various machine learning algorithms such as support vector machines, decision trees, random forests, multilayer perceptrons, and long short-term memories have been examined~\cite{Balahur2014ComparativeEU,Nguyen2018DeepLV,Kumar2020,Rakhmanov:2020a}. \cite{Rakhmanov+Schlippe:2022} showed that transformer models like BERT~\cite{Devlin2019BERTPO} and RoBERTa~\cite{liu2019roberta} (Robustly Optimized BERT Pretraining Approach) generally outperform other machine learning algorithms. Lexicon-based methods, such as those investigated by \cite{Kolchyna2015TwitterSA} and \cite{Kotelnikova:2021}, have also been explored, but machine learning algorithms typically yield better results than lexicon-based approaches. Some researchers suggest using cross-lingual NLP approaches to address the challenges of low-resource languages by leveraging resources from high-resource languages like English~\cite{Vilares:2017,Balahur2014ComparativeEU,Rakhmanov+Schlippe:2022,Lin:2014,Can:2018}. For sentiment analysis, this typically involves translating comments from the original low-resource language into English, enabling the use of well-performing models trained with extensive English resources for the classification task.

Leveraging the pre-trained English BERT model~\cite{Devlin2019BERTPO}, in~\cite{Mabokela+Schlippe_SACAIR:2022} we used cross-lingual sentiment analysis systems to classify tweets in English, Sepedi, and Setswana. To simplify the representation of classified tweet distributions, we defined an \textit{overall sentiment score}, which provides a clear sentiment tendency in a single metric, facilitating topic comparisons. Government institutions can use this score to prioritise and strategically address the issues identified in the tweets. This establishes the foundation for a recommender system that automatically analyses the polarity of text data on the Internet and makes actionable recommendations based on the score. Our AI-driven systems reveal that \textit{employment}, \textit{police service}, \textit{education}, and \textit{health} are particularly problematic for the investigated multilingual communities, with over 50\% of tweets categorised as \textit{negative}, whereas topics like \textit{agriculture} and \textit{rural development} are seen more positively. 

With the advent of LLM-based models like GPT-3.5, GPT-4, LlaMa~2, PaLM~2, and Dolly~2, there is significant potential for their use in sentiment analysis problems. \cite{zhang-etal-2024-sentiment} provide an in-depth investigation into the capabilities of LLM-based chatbots in performing various sentiment analysis tasks, ranging from conventional sentiment classification to aspect-based sentiment analysis and multifaceted analysis of subjective texts---though their study focuses solely on the English language. Their findings indicate that while LLM-based chatbots perform satisfactorily in simpler tasks, they fall short in more complex tasks requiring deeper understanding or structured sentiment information. However, LLM-based chatbots substantially outperform small language models in few-shot learning settings, highlighting their potential when annotation resources are scarce. Furthermore,~\cite{Krugmann2024} evaluate the English sentiment analysis performance of three state-of-the-art LLMs—GPT-3.5, GPT-4, and Llama 2—against established, high-performing transfer learning models. Their research demonstrates that, despite being \textit{zero-shot}, LLMs can not only compete with but also, in some cases, surpass traditional transfer learning methods in sentiment classification accuracy. 

The performance of Africa-centric language models against OpenAI's GPT-3.5 is evaluated by \cite{abbott2023comparing}. They specifically focus on their capabilities in handling low-resourced languages including the Bantu language isiZulu. The study highlights that while ChatGPT and other similar models show impressive results in high-resource languages, their performance significantly drops for African languages due to limited training data and resources. The assessment involves various tasks to demonstrate this disparity and highlights the necessity of developing more inclusive models that cater effectively to underrepresented languages. However---to the best of our knowledge---we are the first to evaluate sentiment analysis of LLMs for South African languages.

\section{Experimental Setup}

In this section, we will first give an overview of our system which determines the degree of action based on the sentiments of the topic-specific social media posts obtained from LLMs. Then, we will present the LLMs which we analysed for sentiment analysis. Furthermore, we will present the prompts we elaborated to instruct the LLMs to classify the social media posts. Finally, we will present the dataset which we used to evaluate the \textit{zero-shot} performance of the analysed LLMs.

\subsection{System Overview}
\begin{figure}
  \centering
  \includegraphics[width=1\linewidth]{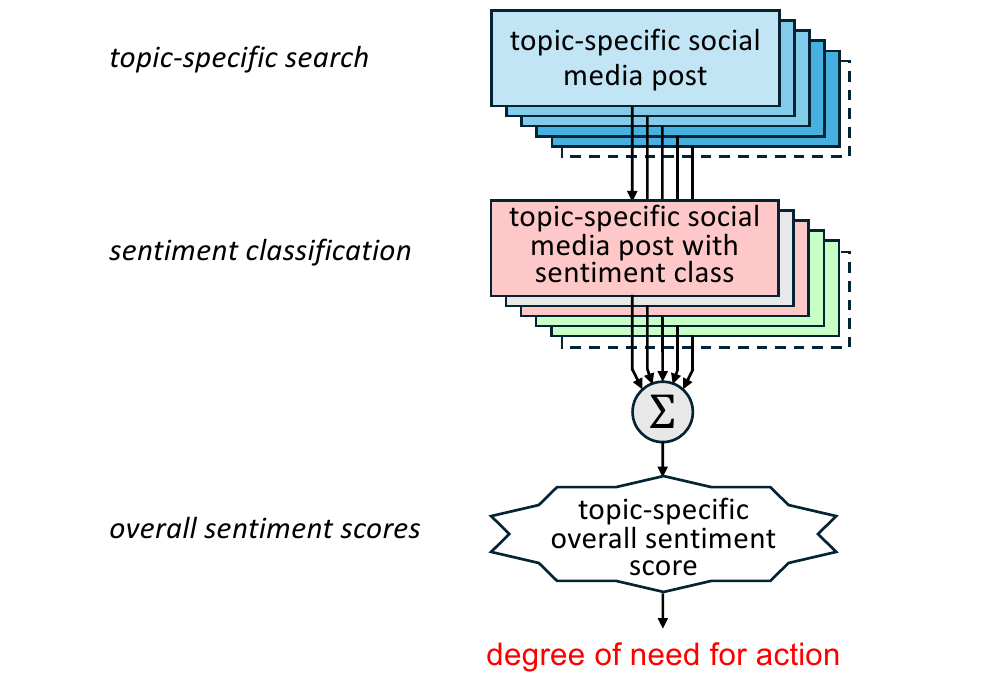}
\caption{Pipeline of Topic-Specific Search, Sentiment Analysis and Scoring.} \label{fig:SystemOverview}
\end{figure}

Figure~\ref{fig:SystemOverview} illustrates the pipeline of our system. Initially, topic-specific social media posts, e.g.~tweets, are gathered using search terms while ensuring data protection measures and applying text normalization steps. Following this, a sentiment analysis system categorises the tweets into \textit{negative}, \textit{neutral}, and \textit{positive} sentiments. Finally, an \textit{overall sentiment score} is calculated for each topic, indicating the degree of need for action.

To represent the distribution of the classified social media posts with only one score, we defined an 

\[overall\ sentiment\ score= \frac{\#positive - \#negative}{\#all sentiments} \]

where $\#positive$ is the number of $positive$ sentiments, $\#negative$ is the number of $negative$ sentiments, and $\#all sentiments$ is the number of all sentiments including the number of elements classified as $\#neutral$.

The \textit{overall sentiment score} ranges from -1 (\textit{completely negative}) to +1 (\textit{completely positive}), offering a clear, single metric to compare topics easily. This score helps governmental institutions prioritize which social challenges to address based on sentiment analysis from tweets. It also forms the basis for a recommender system that analyses text data polarity online and suggests actions based on the sentiment score. The formula can be adjusted to include more sentiment categories or give weight to $neutral$ tweets if needed.

\subsection{Large Language Models}

In this subsection, we will describe the LLMs that we evaluate for sentiment analysis performance on our English, Sepedi and Setswana tweets.

\subsubsection{GPT-3.5}

GPT-3.5 was developed by OpenAI~\cite{baidoo2023education}. The LLM was fine-tuned using reinforcement learning from human feedback~\cite{OpenAI:ChatGPT}. 
This enables the model to understand the meaning and intent behind user inquiries, resulting in relevant and useful responses.  Although OpenAI has not disclosed the specific amount of training data for GPT-3.5, it is known that the prior model, GPT-3, with its 175~billion parameters, was substantially larger than other models such as BERT, RoBERTa, or T5 and was trained on 499~billion tokens~\cite{GPT3:2020}. The LLM can process up to 16k tokens per input~\cite{Nwanne2023}.

\subsubsection{GPT-4}

GPT-4 is available  since March 2023. It was trained on a text corpus of approximately 13~trillion tokens. This text corpus includes well-known sources like \textit{CommonCrawl} and \textit{RefinedWeb}, along with other undisclosed sources~\cite{Patel:2023,Yalalov:2023}.  GPT-4 was first fine-tuned using data from ScaleAI and OpenAI. Subsequently, it was fine-tuned with a reward model (Reinforcement Learning from Human Feedback) and the Proximal Policy Optimization algorithm~\cite{Yalalov:2023,OpenAIGPT4:2023}.
It is estimated that GPT-4 has about 1.8~trillion parameters~\cite{Patel:2023,Yalalov:2023}. The LLM can process up to 128k tokens per input~\cite{Nwanne2023}.

\subsubsection{Dolly 2}

 The open-source LLM Dolly 2.0 was released in April 2023~\cite{zhao2023survey}. Dolly is built on EleutherAI's \textit{pythia} model series \cite{Pythia:2023}. Similar to GPT-3.5, Dolly was fine-tuned to a human-created dataset~\cite{DatabricksBlog2023DollyV2}. The data set contains 15k manually entered entries. Through high-quality fine-tuning, Dolly 2.0 even achieves capabilities that are comparable to GPT-3.5~\cite{DatabricksBlog2023DollyV2}. The LLM can process up to 2k tokens per input~\cite{DatabricksBlog2023DollyV2}.

\subsubsection{PaLM 2}

Google's LLM PaLM~2 was trained with 1.1 trillion parameters~\cite{anil2023palm} and published in May 2023~\cite{anil2023palm}. 
It excels in language comprehension and speech generation, demonstrating outstanding performance in both reasoning and code generation~\cite{PaLM2-Blog}. In the Bison variant which we used, the LLM can process up to 8k tokens per input~\cite{Google:PaLM:2024}.

\subsubsection{LLaMa 2}

The LLaMa 2 model\footnote{https://huggingface.co/meta-llama/Llama-2-70b-chat-hf}, released by Meta in February 2023, features 70 billion parameters. It was fine-tuned for chat instructions using reinforcement learning from human feedback to better align with human preferences for helpfulness and safety. LLaMa 2 outperforms its predecessor, LLaMa 1, which had a maximum of 65 billion parameters~\cite{touvron2023llama}. Additionally, it performs exceptionally well in tests while requiring relatively little computing power~\cite{zhao2023survey}. The LLM can process up to 4k tokens per input~\cite{touvron2023llama}.

\subsubsection{Fusion of the LLM Outputs}

A fusion of machine learning systems' outputs has resulted in better results in other approaches, e.g.~\cite{Agyemang+Schlippe_SACAIR:2024}. Consequently. to get a more precise sentiment classification of the social media posts, we analysed a fusion of the LLM outputs using majority voting. Our idea was to mitigate the misclassification by individual LLMs through this procedure. We counted the classifications for each post and selected the sentiment class chosen by the majority of LLMs as the final class.

\subsection{Prompts for Sentiment Analysis}

\begin{figure}[h]
  \centering
  \includegraphics[width=1\linewidth]{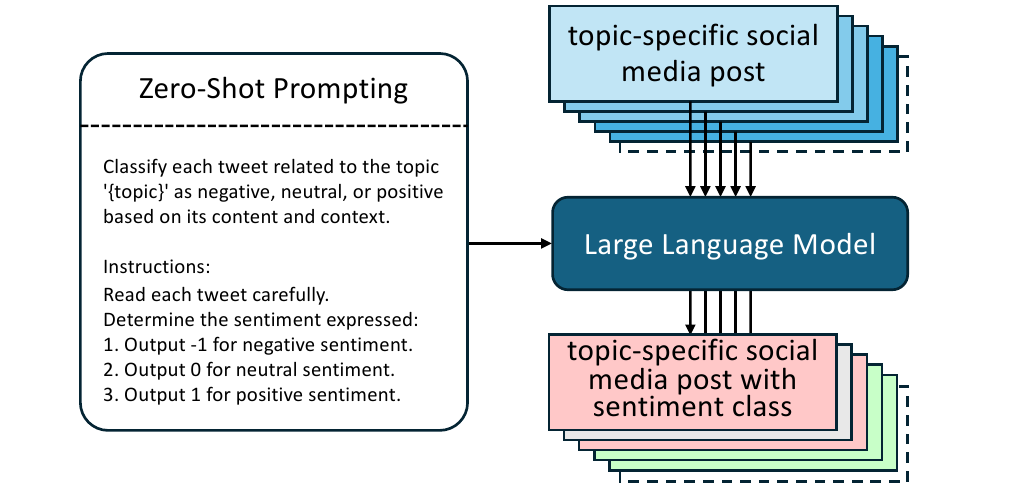}
\caption{\textit{Zero-Shot} Sentiment Classification Workflow with Prompting Example and Expected Response from the LLMs.} \label{fig:prompting}
\end{figure}

Figure~\ref{fig:prompting} presents the prompts we elaborated to instruct the LLMs to classify the social media posts. For prompt engineering, it was important for us to pass a precise description to the LLMs. Consequently, for each topic, we used a separate prompt, where we added the instruction to classify each social media post, i.e.~tweet in our case, indicated the topic and defined the three sentiment classes $negative$, $neutral$ and $positive$. Since we detected that our analysed LLMs can handle tables in csv format well, we added the list of tweets belonging to the corresponding topic in csv format. Since our initial experiments with GPT-4 demonstrated that English prompts lead to better results than prompts in the native language---1.0\% relative better F1 scores for Sepedi and 1.5\% for Setswana---we decided to use English prompts. Similar findings concerning English vs.~foreign prompts are reported in~\cite{kmainasi2024nativevsnonnativelanguage}. Note that the topic classification or the text classification, which we did manually, could also be done by the LLMs as shown in~\cite{10367969,sun-etal-2023-text}.

\subsection{The SAGovTopicTweets Corpus}
Our goal was to analyse state-of-the-art LLMs’ sentiment analysis performances to detect social challenges. Consequently, we used the \textit{SAGovTopicTweets Corpus} for our experiments which we had specifically collected for this use case as described in~\cite{Mabokela+Schlippe_SACAIR:2022}. The \textit{SAGovTopicTweets Corpus} contains South African tweets in English, Setswana, and Sepedi covering the topics \textit{employment}, \textit{sanitation}, \textit{police service}, \textit{education}, \textit{health}, \textit{small business}, \textit{transport}, \textit{home affair}, \textit{rural development}, and \textit{agriculture}---10~government departments related topics that were highlighted in the State of the Nation Address for 2021 as key government issues to strengthen the economy~\cite{SONA2022}. 
Of course, English, Setswana, and Sepedi are a subset of the languages spoken in South Africa. But in the other datasets,  the topics are not annotated. Nevertheless, a proof-of-concept can be achieved with this subset. 
The \textit{SAGovTopicTweets Corpus} contains 16,787~tweets across languages and topics. The topics are equally distributed over the languages as described in~\cite{Mabokela+Schlippe_SACAIR:2022}. The average number of word tokens per tweet is 21 for English, 12 for Sepedi, and 11 for Setswana, i.e. 15 on average across all three languages.

\section{Experiments and Results}

By employing AI for sentiment analysis, governments and other stakeholders can gain insights into the expressions of feelings and attitudes among diverse communities, which can assist in identifying and addressing social issues. Proactive analysis can facilitate timely interventions by policymakers, healthcare providers, and social workers, which can ultimately contribute to societal well-being. The accurate classification of sentiment is of paramount importance in these tasks, as it ensures a precise understanding of public emotions and reactions. Accordingly, this section presents a comprehensive technical evaluation and a socio-cultural interpretation of the sentiment analysis data. Our technical evaluation is designed to assess the methodology, accuracy, and reliability of the sentiment analysis across different languages and topics. Our socio-cultural interpretation seeks to contextualize the sentiment variations by investigating social nuances expressed in the different languages. By integrating these perspectives, our objective is to provide a comprehensive understanding of the data, emphasizing both the technological robustness and the cultural relevance of the findings.

\subsection{Sentiment Classification Performances of the LLMs}

To have a system which determines the degree of action based on the sentiments of the government-related topics, it is important to (1) have an excellent sentiment classification performance for all topics~\cite{bhatia-p-2018-topic}, (2) have an excellent sentiment classification performance for all languages so that all language groups are well represented~\cite{VILARES2017595}. Consequently, our goal was to analyze these two dimensions. 

To evaluate the sentiment classification performance of the different LLMs, we conducted a study to determine their misclassification rates across different languages and topics. The results of this study are presented in Table~\ref{tab:sentimentClassification}. A lower value indicates less classification errors in comparison to the human-evaluated reference. 

Looking at the error rates in sentiment classification, GPT-3.5 generally exhibits higher sentiment errors. GPT-4 tends to show the lowest sentiment errors across all topics (6.5\%--10.9\%). LLaMa~2 (9.7\%--13.0\%) and Dolly~2 (10.0\%--12.1\%) are relatively similar, often between GPT-3.5 (10.0\%--13.8\%) and GPT-4 (6.3\%--10.9\%). PaLM 2 (6.5\%--12.6\%) provides the second-best overall performance, right after GPT-4. 
According to the independent samples t-test, the overall errors in sentiment classification for Dolly~2~(11.6\%) show statistical equivalence with both LLaMa~2~(11.5\%) and GPT-3.5~(12.5\%), indicating similar performance levels. This suggests that, despite the differences in error rates in sentiment classification per topic, the overall sentiment analysis capabilities of Dolly~2 are comparable to LLaMa~2 and GPT-3.5. 

The majority voting approach in the fused system leads to lower error rates (0.2\%-0.9\%) for all LLMs by providing a more reliable, stable, and balanced sentiment classification, making it ideal for applications requiring consistency and robustness.


\begin{table}[!ht]
    \centering
    \caption{LLMs' error rates in sentiment classification across topics and languages. Note: All annotators disagree on a subset of 1k posts in 0.6\% of the tweets.}
    \label{tab:sentimentClassification}
    \begin{tabular}{lrrrrrr}
    \hline
        \textbf{} & \textbf{~~GPT-3.5} & \textbf{~~GPT-4} & \textbf{~~LLaMa 2} & \textbf{~~PaLM 2} & \textbf{~~Dolly 2} & \textbf{~~Fused} \\ \hline
        agriculture & 11.2\% & 6.5\% & 10.9\% & 8.4\% & 11.9\% & 0.3\%  \\
        education & 13.0\% & 8.4\% & 9.9\% & 8.9\% & 12.1\% & 0.5\%  \\
        employment & 10.6\% & 6.7\% & 10.0\% & 6.5\% & 10.3\% & 0.3\%  \\
        health & 13.5\% & 8.5\% & 11.0\% & 8.7\% & 12.5\% & 0.2\%  \\
        home affairs & 12.4\% & 8.6\% & 12.7\% & 10.3\% & 12.1\% & 0.4\%  \\
        police service & 12.9\% & 9.0\% & 12.6\% & 10.0\% & 11.0\% & 0.9\%  \\
        rural development & 13.8\% & 6.3\% & 10.5\% & 12.6\% & 11.9\% & 0.3\%  \\
        sanitation & 12.5\% & 7.0\% & 11.5\% & 8.9\% & 11.3\% & 0.6\%  \\
        small business & 12.6\% & 7.5\% & 13.0\% & 10.4\% & 11.2\% & 0.6\%  \\
        transport & 12.5\% & 10.9\% & 11.7\% & 8.8\% & 12.1\% & 0.6\%  \\ \hline
        English & 12.8\% & 8.6\% & 11.9\% & 9.5\% & 12.0\% & 0.4\%  \\
        Sepedi & 12.3\% & 7.0\% & 9.7\% & 8.0\% & 10.0\% & 0.7\%  \\
        Setswana & 10.0\% & 7.3\% & 12.2\% & 8.8\% & 11.8\% & 0.6\%  \\ \hline
        Overall & 12.5\% & 8.2\% & 11.5\% & 9.2\% & 11.6\% & 0.5\% \\ \hline
    \end{tabular}
\end{table}


Comparing the effectiveness of fusing sentiment classifications from different systems over the three languages demonstrates that the less similar the systems are, i.e.~the lower their correlation, the more effective their fusion:  
English benefits the most (best: 8.6\%, fusion: 0.4\%) from the fusion of sentiment results due to the lowest correlation\footnote{The correlations are calculated using Pearson's r.} between the classified sentiments of 0.770 among individual LLMs in comparison to 0.792 for Sepedi and 0.803 for Setswana. 
This low correlation indicates significant differences in the LLMs' outputs, which fusion helps to average out, leading to a more stable and reliable sentiment score.

Sepedi shows the lowest fusion gain (best: 7.0\%, fusion: 0.4\%) due to lower variance and higher correlation between the LLMs, meaning that individual LLMs are more equal in their outputs. 
Setswana has a medium level of variance and correlation, leading to moderate gains from the fusion process (best: 7.3\%, fusion: 0.4\%)

Even though it can be assumed that English 
has more training data to train the LLMs due to a much higher number of speakers (380 million native speakers) than Sepedi (4.7 million native speakers) and Setswana (6.6 million native speakers)~\cite{SACensus2022}, our sentiment analysis provides robust results for all three investigated languages.

\begin{table}[!ht]
    \centering
   \caption{LLMs' F1-scores in sentiment classification across languages}
\label{tab:sentimentClassification:F1}
    \begin{tabular}{lrrrrrr}
    \hline
        \textbf{} & \textbf{~~GPT-3.5} & \textbf{~~GPT-4} & \textbf{~~LLaMa 2} & \textbf{~~PaLM 2} & \textbf{~~Dolly 2} & \textbf{~~Fused} \\ 
        \hline
English   & 91.0\% & 94.1\% & 91.6\% & 93.2\% & 91.5\% & 97.2\% \\
Sepedi    & 91.9\% & 95.3\% & 93.6\% & 94.5\% & 93.3\% & 97.8\% \\
Setswana  & 93.2\% & 95.3\% & 92.4\% & 94.3\% & 92.5\% & 97.4\% \\
\hline
Overall   & 91.4\% & 94.4\% & 92.0\% & 93.6\% & 91.9\% & 97.5\% \\
\hline

\hline
    \end{tabular}
\end{table}

For comparison to other studies, we have also listed the LLMs' F1 scores in Table~\ref{tab:sentimentClassification:F1}. We see that the LLMs' performance is significantly better than what was reported in~\cite{Mabokela+Schlippe_SACAIR:2022} on the SAfriSenti test sets: The BERT-based English system had an F1 score of 86.0\%, the Sepedi system had an F1 score of 84.0\%, and the Setswana system had an F1 score of 82.7\%. This shows that the state-of-the-art LLMs can obtain better sentiment analysis performances than more traditional deep learning-based approaches.

\subsection{Socio-Cultural Interpretation}

\begin{figure}[!h]
  \centering
  \includegraphics[width=1\linewidth]{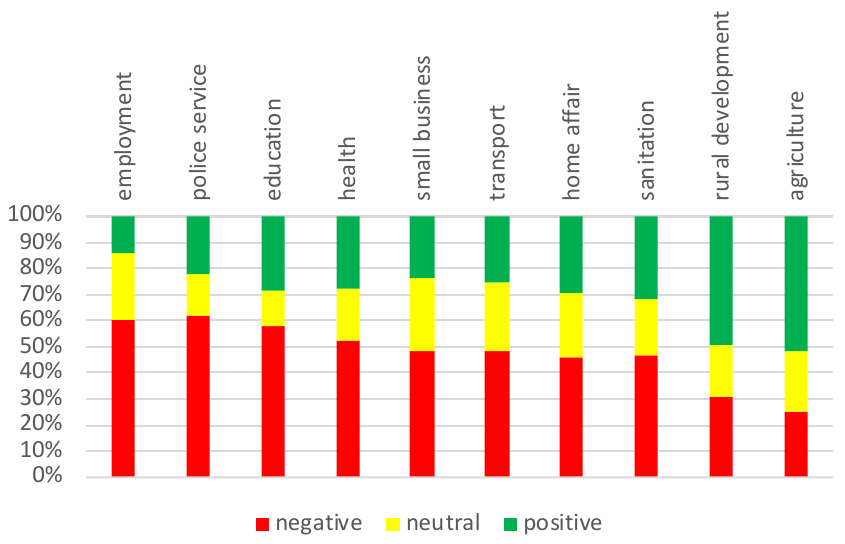}
\caption{Sentiment distribution of the investigated topics.} \label{fig:SentimentDistribution}
\end{figure}

After demonstrating the excellent sentiment
classification performance for all topics and languages---particularly with the LLMs' fusion, we conducted a socio-cultural interpretation of the sentiment distributions.  
Figure~\ref{fig:SentimentDistribution} shows \textit{negative}, \textit{neutral} and \textit{positive} sentiment distributions per topic. The distributions indicate that the topics \textit{employment}, \textit{police service}, \textit{education}, and \textit{health} are particularly problematic, as more than 50\% of the tweets are scored $negative$. The sentiments regarding \textit{agriculture} and \textit{rural development} is rather \textit{positive}. 

\begin{figure}[h!]
  \centering
  \includegraphics[width=1\linewidth]{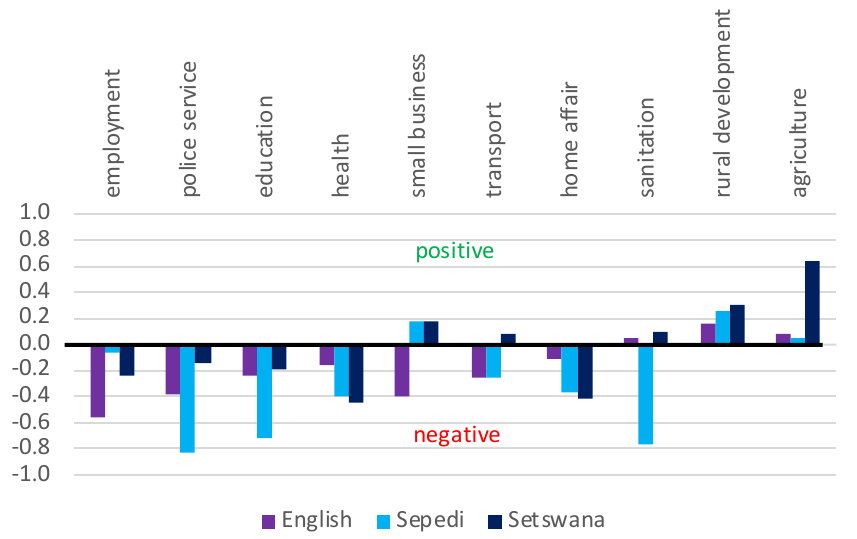}
\caption{\textit{Overall sentiment score} per language.} \label{fig:SentimentPerLanguage}
\end{figure}

A better overview of the languages is visualised in the \textit{overall sentiment scores} in Figure~\ref{fig:SentimentPerLanguage}. The figure reveals significant differences in how topics are perceived across different languages. Setswana tends to have more $positive$ \textit{overall sentiment scores} (e.g., for \textit{rural development} 0.30 and \textit{agriculture} 0.64, on average the \textit{overall sentiment score} over all topics is -0.01), while English and Sepedi exhibit a more $neutral$ \textit{overall sentiment score} (on average -0.18) or slightly $negative$ \textit{overall sentiment scores} over the topics (on average -0.29) in comparison. These variations could be influenced by cultural, socio-economic, and linguistic factors that shape how individuals express their views and opinions on different topics. If the recommender system identifies a $negative$ sentiment towards a particular topics, the government could implement community-based initiatives to improve service delivery specifically in the area concerned. This shows again the significance of reporting the sentiment of the individual languages and cultures.

Investigating individual topics and their relative change in \textit{overall sentiment scores} when comparing Sepedi and Setswana to English, we observe that Sepedi provides significantly lower topic-specific \textit{overall sentiment scores} in particular for \textit{police service} (-0.84), \textit{education} (-0.71), and \textit{sanitation} (-0.77). Sepedi and English speakers may have different cultural norms and expectations regarding these services. Sepedi speakers may be more critical or have different standards for what constitutes good service. Both English and Sepedi show a move towards neutrality compared to Setswana's $positive$ sentiment for \textit{agriculture} (0.64). For \textit{employment} English shows a particular low \textit{overall sentiment score} (-0.55) in contrast to the other two languages.

\subsection{Discussion}

Our results demonstrate the usability and strong performance of the tested state-of-the-art LLMs for sentiment analysis on social media data to assess the level of need for action on social issues. For example, the F1 scores show that the LLMs’ performance in sentiment analysis is comparable to other advanced systems. We would now like to briefly address some limitations that should be considered in future work to enable concrete use in a recommender system that identifies social issues.

Our analyses show that combining the outputs of multiple LLMs significantly reduces sentiment classification errors. However, this improvement in quality comes at the cost of increased computational resources, presenting a tradeoff between cost and performance. In our view, the gains in quality justify the added expense. 
Additionally, the analysis relies on the availability of relevant social media posts on the selected platforms, which must be accurately found and collected. To ensure realistic assessments for a given period, the posts must match the specified timeframe. Furthermore, new social media posts need to be continuously classified, although some search engines already offer features to search for specific topics. 
Finally, the concrete actions taken in response to the identified need for action should be defined by the responsible bodies based on the insights provided by our system.

\vspace{-0.2cm}

\section{Conclusion and Future Work}

In conclusion, this study demonstrated the potential of state-of-the-art LLMs in addressing social challenges in South Africa by performing sentiment analysis on social media posts in English, Sepedi, and Setswana and by providing the degree of action needed in form of an \textit{overall sentiment score}. By leveraging our SAGovTopicTweets corpus, which covers key topics related to South African government departments, the study evaluated the performance of various LLMs. 

We found out that combining the outputs from our different LLMs significantly enhances sentiment classification performance, achieving sentiment classification errors below 1\%. Consequently, it is now feasible to develop systems for English, Sepedi and Setswana that reliably generate sentiment analysis information to detect social challenges and inform necessary actions across different topics and language groups using LLMs.

However, based on the diversity of the resulting \textit{overall sentiment scores} in the topics and languages, we learned that it is important to check the sentiment for each topic and language instead of looking at them in general. The reason can be due to social environment or that different languages often express sentiments in unique ways, influenced by cultural nuances, vocabulary, and syntax.

Therefore, future work must investigate if more pronounced $negative$ or $positive$ sentiments between the languages for the same topics are due to linguistic and cultural differences or since the community is underserved.

\vspace{-0.2cm}

%
%

%
%
 \bibliographystyle{splncs04}
 \bibliography{mybibliography}

@article{Kaur:2020,
author = {Kaur, Chhinder and Sharma, Anand},
year = {2020},
month = {08},
pages = {6303-6311},
title = {{Sentiment Analysis of Tweets on Social Issues using Machine Learning Approach}},
volume = {9},
journal = {International Journal of Advanced Trends in Computer Science and Engineering},
doi = {10.30534/ijatcse/2020/310942020      }
}

@inproceedings{mabokela-etal-2023-investigating,
    title = {{Investigating Sentiment-Bearing Words- and Emoji-based Distant Supervision Approaches for Sentiment Analysis}},
    author = "Mabokela, Ronny  and
      Roborife, Mpho  and
      Celik, Turguy",
    editor = "Mabuya, Rooweither  and
      Mthobela, Don  and
      Setaka, Mmasibidi  and
      Van Zaanen, Menno",
    booktitle = "Proceedings of the Fourth workshop on Resources for African Indigenous Languages (RAIL 2023)",
    month = may,
    year = "2023",
    address = "Dubrovnik, Croatia",
    publisher = "Association for Computational Linguistics",
    url = "https://aclanthology.org/2023.rail-1.13",
    doi = "10.18653/v1/2023.rail-1.13",
    pages = "115--125",
}

@inproceedings{bhatia-p-2018-topic,
    title = "Topic-Specific Sentiment Analysis Can Help Identify Political Ideology",
    author = "Bhatia, Sumit  and
      P, Deepak",
    editor = "Balahur, Alexandra  and
      Mohammad, Saif M.  and
      Hoste, Veronique  and
      Klinger, Roman",
    booktitle = "Proceedings of the 9th Workshop on Computational Approaches to Subjectivity, Sentiment and Social Media Analysis",
    month = oct,
    year = "2018",
    address = "Brussels, Belgium",
    publisher = "Association for Computational Linguistics",
    url = "https://aclanthology.org/W18-6212",
    doi = "10.18653/v1/W18-6212",
    pages = "79--84",
}

@article{VILARES2017595,
title = {Supervised sentiment analysis in multilingual environments},
journal = {Information Processing \& Management},
volume = {53},
number = {3},
pages = {595-607},
year = {2017},
issn = {0306-4573},
doi = {https://doi.org/10.1016/j.ipm.2017.01.004}    ,
url = {https://www.sciencedirect.com/science/article/pii/S0306457316302540},
author = {David Vilares and Miguel A. Alonso and Carlos Gómez-Rodríguez}
}

@misc{SACensus2022,
  TITLE =         {{South African Population}},
  INSTITUTION =   {Statistics South Africa}, 
  YEAR  =         {2022},
  URL = {https://census.statssa.gov.za/\#/}
}

@InProceedings{Mabokela+Schlippe_SACAIR:2022,
author="Mabokela, Koena Ronny
and Schlippe, Tim",
editor="Pillay, Anban
and Jembere, Edgar
and Gerber, Aurona",
title={{AI for Social Good: Sentiment Analysis to Detect Social Challenges in South Africa}},
booktitle="Artificial Intelligence Research",
year="2022",
publisher="Springer Nature Switzerland",
address="Cham",
pages="309--322",
isbn="978-3-031-22321-1"
}

@InProceedings{Agyemang+Schlippe_SACAIR:2024,
author="Agyemang, Alex
and Schlippe, Tim",
title={{AI in Education: An Analysis of Large Language Models for Twi Automatic Short Answer Grading}},
booktitle="Artificial Intelligence Research",
year="2024",
publisher="Springer Nature Switzerland",
address="Cham"
}

@misc{kmainasi2024nativevsnonnativelanguage,
      title={{Native vs Non-Native Language Prompting: A Comparative Analysis}}, 
      author={Mohamed Bayan Kmainasi and Rakif Khan and Ali Ezzat Shahroor and Boushra Bendou and Maram Hasanain and Firoj Alam},
      year={2024},
    eprint={2409.07054},
  archivePrefix={arXiv},
    primaryClass={cs.CL},
      url={https://arxiv.org/abs/2409.07054}, 
}

@INPROCEEDINGS{Makuwe:2023,
  author={Makuwe, Barlette and Mabokela, Koena Ronny and Schlippe, Tim},
  booktitle={11th International Conference on Affective Computing and Intelligent Interaction (ACII)}, 
  title={{Sentiment Analysis for Shona}}, 
  year={2023},
  volume={},
  number={},
  pages={1-8},
  doi={10.1109/ACII59096.2023.10388095}}

@article{Krugmann2024,
  title={Sentiment Analysis in the Age of Generative AI},
  author={Krugmann, J.O. and Hartmann, J.},
  journal={Customer Needs and Solutions},
  volume={11},
  number={3},
  year={2024},
  doi={10.1007/s40547-024-00143-4},
  url={https://doi.org/10.1007/s40547-024-00143-4}       
}

@inproceedings{zhang-etal-2024-sentiment,
    title = {{Sentiment Analysis in the Era of Large Language Models: A Reality Check}},
    author = "Zhang, Wenxuan  and
      Deng, Yue  and
      Liu, Bing  and
      Pan, Sinno  and
      Bing, Lidong",
    editor = "Duh, Kevin  and
      Gomez, Helena  and
      Bethard, Steven",
    booktitle = "Findings of the Association for Computational Linguistics: NAACL 2024",
    month = jun,
    year = "2024",
    address = "Mexico City, Mexico",
    publisher = "Association for Computational Linguistics",
    url = "https://aclanthology.org/2024.findings-naacl.246",
    pages = "3881--3906"
}

@TECHREPORT{CountryReportSouthAfrica,
  TITLE =         {{Sustainable Development Goals: Country Report 2019 -- South Afrcia}},
  NUMBER =        {ISBN 978-0-621-47619-4 },
  INSTITUTION =   {Statistics South Africa}, 
  YEAR  =         {2019}
}

@inproceedings{Devlin2019BERTPO,

  title={{BERT: Pre-training of Deep Bidirectional Transformers for Language Understanding}},
  author={Jacob Devlin and Ming-Wei Chang and Kenton Lee and Kristina Toutanova},
  booktitle={NAACL},
  year={2019}
}

@article{Musikanski2020ArtificialIA,
  title={{Artificial Intelligence and Community Well-being: A Proposal for an Emerging Area of Research}},
  author={Laura Musikanski and Bogdana Rakova and James Bradbury and Rhonda G. Phillips and Margaret Manson},
  journal={International Journal of Community Well-Being},
  year={2020},
  volume={3},
  pages={39-55}
}

@article{Hager2019ArtificialIF,
  title={{Artificial Intelligence for Social Good}},
  author={Gregory Hager and Ann W. Drobnis and Fei Fang and Rayid Ghani and Amy Greenwald and Terah Lyons and David C. Parkes and Jason Schultz and Suchi Saria and Stephen F. Smith and Milind Tambe},
  journal={ArXiv},
  year={2019},
  volume={abs/1901.05406}
}

@article{Tomaev2020AIFS,
  title={{AI for Social Good: Unlocking the Opportunity for Positive Impact}},
  author={Nenad Tomasev and Julien Cornebise and Frank Hutter and S. Mohamed and Angela Picciariello and Bec Connelly and Danielle Belgrave and Daphne Ezer and Fanny Cachat van der Haert and Frank Mugisha and Gerald Abila and Hiromi Arai and Hisham Almiraat and Julia Proskurnia and Kyle Snyder and Mihoko Otake-Matsuura and Mustafa Farooq Othman and Tobias Glasmachers and Wilfried de Wever and Yee Whye Teh and Mohammad Emtiyaz Khan and Ruben De Winne and Tom Schaul and Claudia Clopath},
  journal={Nature Communications},
  year={2020},
  volume={11}
}

@article{Corneliu2022,
author = {Corneliu Bjola},
title = {{AI for Development: Implications for Theory and Practice}},
journal = {Oxford Development Studies},
volume = {50},
number = {1},
pages = {78-90},
year  = {2022},
publisher = {Routledge},
doi = {10.1080/13600818.2021.1960960   },

URL = {https://doi.org/10.1080/13600818.2021.1960960   },
eprint = {https://doi.org/10.1080/13600818.2021.1960960   }

}

@article{Wankhade:2022,
author = {Wankhade, Mayur and Rao, Annavarapu and Kulkarni, Chaitanya},
year = {2022},
month = {02},
pages = {1--50},
title = {{A Survey on Sentiment Analysis Methods, Applications, and Challenges}},
journal = {Artificial Intelligence Review},
doi = {10.1007/s10462-022-10144-1  }
}

@inproceedings{nakov2016semeval,
	title={{SemEval-2016 Task 4: Sentiment Analysis in Twitter}},
	author={Nakov, Preslav and Ritter, Alan and Rosenthal, Sara and Sebastiani, Fabrizio and Stoyanov, Veselin},
	booktitle={International Workshop on Semantic Evaluation (SemEval)},
	year={2016}
}

@inproceedings{paktwitter2010,
	title={{Twitter as a Corpus for Sentiment Analysis and Opinion Mining}},
	author={Pak, Alexander and Paroubek, Patrick},
	booktitle={The 7th Edition of the Language Resources and Evaluation Conference (LREC 2010)},
	pages={1320--1326},
	year={2010}
}

@inproceedings{Agarwal2012EndtoEndSA,
  title={{End-to-End Sentiment Analysis of Twitter Data}},
  author={Apoorv Agarwal and Jasneet Singh Sabharwal},
  booktitle={Conference: Proceedings of the Workshop on Information Extraction and Entity Analytics on Social Media Data},
  year={2012}
}

@misc{Statista:2022,
  title = {{Statista: The Most Spoken Languages Worldwide in 2022}},
  howpublished = {\url{https://www.statista.com/statistics/266808/the-most-spoken-languages-worldwide}},
  year = 2022,
  note = {Accessed: 08-2022}
}

@misc{SDGs:2022,
  author = {{United Nations}},
  title = {{Sustainable Development Goals: 17 Goals to Transform our World}},
  howpublished = {https://www.un.org/sustainabledevelopment/sustainabledevelopment-goals},
  note = {Accessed: 2022-08},
  year = 2022
}

@article{Balahur2014ComparativeEU,
  title={{Comparative Experiments using Supervised Learning and Machine Translation for Multilingual Sentiment Analysis}},
  author={Alexandra Balahur and Marco Turchi},
  journal={Comput. Speech Lang.},
  year={2014},
  volume={28},
  pages={56-75}
}

@article{Nguyen2018DeepLV,
  title={{Deep Learning versus Traditional Classifiers on Vietnamese Students’ Feedback Corpus}},
  author={Phu X. V. Nguyen and Tham Vo Thi Hong and Kiet Van Nguyen and Ngan Luu-Thuy Nguyen},
  journal={The 5th NAFOSTED Conference on Information and Computer Science (NICS)},
  year={2018}
}

@Inbook{Kumar2020,
author="Kumar, Ashish
and Sharan, Aditi",
editor="Agarwal, Basant
and Nayak, Richi
and Mittal, Namita
and Patnaik, Srikanta",
title="Deep Learning-Based Frameworks for Aspect-Based Sentiment Analysis",
bookTitle="Deep Learning-Based Approaches for Sentiment Analysis",
year="2020",
publisher="Springer Singapore",
pages="139--158",
isbn="978-981-15-1216-2"
}

@article{Rakhmanov:2020a,
author = {Rakhmanov, Ochilbek},
year = {2020},
pages = {194-204},
title = {{A Comparative Study on Vectorization and Classification Techniques in Sentiment Analysis to Classify Student-Lecturer Comments}},
volume = {178},
journal = {Procedia Computer Science}
}

@article{Kolchyna2015TwitterSA,
  title={{Twitter Sentiment Analysis: Lexicon Method, Machine Learning Method and Their Combination}},
  author={Olga Kolchyna and Th{\'a}rsis Tuani Pinto Souza and Philip C. Treleaven and Tomaso Aste},
  journal={arXiv: Computation and Language},
  year={2015}
}

@inproceedings{Kotelnikova:2021,
author = {Kotelnikova, Anastasia and Paschenko, Danil and Bochenina, Klavdiya and Kotelnikov, Evgeny},
year = {2021},
title = {{Lexicon-based Methods vs. BERT for Text Sentiment Analysis}},
booktitle = {AIST}
}

@INPROCEEDINGS{Lin:2014,
  author={Lin, Zheng and Jin, Xiaolong and Xu, Xueke and Wang, Yuanzhuo and Tan, Songbo and Cheng, Xueqi},
  booktitle={IEEE/WIC/ACM International Joint Conferences on Web Intelligence (WI) and Intelligent Agent Technologies (IAT)}, 
  title={{Make It Possible: Multilingual Sentiment Analysis Without Much Prior Knowledge}}, 
  year={2014},
  volume={2},
  number={},
  pages={79--86},
  doi={10.1109/WI- IAT.2014.83}}

@article{Vilares:2017,
author = {Vilares, David and Alonso Pardo, Miguel and Gómez-Rodríguez, Carlos},
year = {2017},
month = {05},
pages = {},
title = {{Supervised Sentiment Analysis in Multilingual Environments}},
volume = {53},
journal = {Information Processing \& Management},
doi = {10.1016/j. ipm.2017.01.004}
}

@INPROCEEDINGS{Can:2018,
author    =  {Ethem F. Can and
               Aysu Ezen{-}Can and
               Fazli Can},
  booktitle={ACM SIGIR 2018 Workshop on Learning from Limited or Noisy Data}, 
  title={{Multilingual Sentiment Analysis: An RNN-Based Framework for Limited Data}}, 
  year={2018},
}

@inproceedings{Rakhmanov+Schlippe:2022,
author = {Ochilbek Rakhmanov and Tim Schlippe},
year = {2022},
title = {{Sentiment Analysis for Hausa: Classifying Students' Comments}},
booktitle = {The 1st Annual Meeting of the ELRA/ISCA Special Interest Group on Under-Resourced Languages (SIGUL 2022)},
  address   = "Marseille, France"
}

@misc{liu2019roberta,
      title={{RoBERTa: A Robustly Optimized BERT Pretraining Approach}}, 
      author={Yinhan Liu and Myle Ott and Naman Goyal and Jingfei Du and Mandar Joshi and Danqi Chen and Omer Levy and Mike Lewis and Luke Zettlemoyer and Veselin Stoyanov},
      year={2019},
      eprint={1907.11692},
      archivePrefix={arXiv},
      primaryClass={cs.CL}
}

@article{Go2009,
author = {Go, Alec and Bhayani, Richa and Huang, Lei},
year = {2009},
month = {01},
pages = {},
title = {{Twitter Sentiment Classification using Distant Supervision}},
volume = {150},
journal = {Processing}
}

@article{Indriani2020,
  author    = {Dian Indriani and
               Arbi Haza Nasution and
               Winda Monika and
               Salhazan Nasution},
  title     = {{Towards a Sentiment Analyzer for Low-Resource Languages}},
  journal   = {CoRR},
  volume    = {abs/2011.06382},
  year      = {2020},
  eprinttype = {arXiv}
}

@article{AIU4SG:2020,
  author    = {Zheyuan Ryan Shi and
               Claire Wang and
               Fei Fang},
  title     = {{Artificial Intelligence for Social Good: A Survey}},
  journal   = {CoRR},
  volume    = {abs/2001.01818},
  year      = {2020}
}

@article{Kiritchenko2018ExaminingGA,
  title={{Examining Gender and Race Bias in Two Hundred Sentiment Analysis Systems}},
  author={Svetlana Kiritchenko and Saif M. Mohammad},
  journal={ArXiv},
  year={2018},
  volume={abs/1805.04508}
}

@misc{SONA2022,
    author = {Cyril Ramaphosa},
    title = {{State of the Nation Address}},
    URL = {https://www.stateofthenation.gov.za/assets/2021/SONA\%202021.pdf},
    note = {accessed: 08-2022},
    year={2021}
}

@article{baidoo2023education,
  title={{Education in the Era of Generative Artificial Intelligence (AI): Understanding the Potential Benefits of ChatGPT in Promoting Teaching and Learning}},
  author={Baidoo-Anu, David and Owusu Ansah, Leticia},
  journal={SSRN 4337484},
  year={2023}
}

@ARTICLE{10367969,
  author={Fatemi, Bahareh and Rabbi, Fazle and Opdahl, Andreas L.},
  journal={IEEE Access}, 
  title={{Evaluating the Effectiveness of GPT Large Language Model for News Classification in the IPTC News Ontology}}, 
  year={2023},
  volume={11},
  number={},
  pages={145386-145394},
  doi={10.1109/ACCESS.2023.3345414}
}

@inproceedings{sun-etal-2023-text,
    title = {{Text Classification via Large Language Models}},
    author = "Sun, Xiaofei  and
      Li, Xiaoya  and
      Li, Jiwei  and
      Wu, Fei  and
      Guo, Shangwei  and
      Zhang, Tianwei  and
      Wang, Guoyin",
    editor = "Bouamor, Houda  and
      Pino, Juan  and
      Bali, Kalika",
    booktitle = "Findings of the Association for Computational Linguistics: EMNLP 2023",
    month = dec,
    year = "2023",
    address = "Singapore",
    publisher = "Association for Computational Linguistics",
    url = "https://aclanthology.org/2023.findings-emnlp.603",
    doi = "10.18653/v1/2023.findings-emnlp.603",
    pages = "8990--9005"
}

@misc{abbott2023comparing,
  title = {{Comparing Africa-centric Models to OpenAI’s GPT-3.5}},
  author = {Jade Abbott and Bonaventure Dossou and Rooweither Mbuya},
  year = {2023},
  howpublished = {\url{https://lelapa.ai/comparing-africa-centric-models-to-openais-gpt3-5-2}},
  note = {Lelapa AI, Accessed: 2024-07-29}
}

@misc{OpenAI:ChatGPT,
  author = {OpenAI},
  title = {{What is ChatGPT?}},
  year = {2023},
  url = {https://help.openai.com/en/articles/6783457-what-is-chatgpt},
  urldate = {2023-04-14}
}

@article{GPT3:2020,
  author       = {Tom B. Brown and
                  Benjamin Mann and
                  Nick Ryder and
                  Melanie Subbiah and
                  Jared Kaplan and
                  Prafulla Dhariwal and
                  Arvind Neelakantan and
                  Pranav Shyam and
                  Girish Sastry and
                  Amanda Askell and
                  Sandhini Agarwal and
                  Ariel Herbert{-}Voss and
                  Gretchen Krueger and
                  Tom Henighan and
                  Rewon Child and
                  Aditya Ramesh and
                  Daniel M. Ziegler and
                  Jeffrey Wu and
                  Clemens Winter and
                  Christopher Hesse and
                  Mark Chen and
                  Eric Sigler and
                  Mateusz Litwin and
                  Scott Gray and
                  Benjamin Chess and
                  Jack Clark and
                  Christopher Berner and
                  Sam McCandlish and
                  Alec Radford and
                  Ilya Sutskever and
                  Dario Amodei},
  title        = {{Language Models are Few-Shot Learners}},
  journal      = {CoRR},
  volume       = {abs/2005.14165},
  year         = {2020},
  eprinttype    = {arXiv}
}

@misc{Patel:2023,
    author = {Patel, Dylan and Wong, Gerald},
    title = {{GPT-4 Architecture, Infrastructure, Training Dataset, Costs, Vision, MoE}},
    year = {2023},
    month = {July},
    publisher = {GitHub},
    journal = {GitHub repository},
    howpublished = {https://github.com/llv22/gpt4\_essay/blob/master/GPT-4-4.JPG},
    note = {Accessed: 30-09-2023}
}

@article{Yalalov:2023,
    title = {{GPT-4's Leaked Details Shed Light on its Massive Scale and Impressive Architecture}},
    author = {Yalalov, Damir and Myakin, Danil},
    year = {2023},
    month = {July},
    journal = {Metaverse Post},
    url = {https://mpost.io/gpt-4s-leaked-details-shed-light-on-its-massive-scale-and-impressive-architecture/\#gpt-4s-massive-parameters-count}
}

@misc{OpenAIGPT4:2023,
    title = {{GPT-4}},
    author = {OpenAI},
    year = {2023},
    month = {March},
    journal = {OpenAI Research},
    url = {https://openai.com/research/gpt-4}
}

@article{Nwanne2023,
    author = {Winnie Nwanne},
    title = {Comparing GPT-3.5 \& GPT-4: A Thought Framework on When to Use},
    year = {2023},
    journal = {AI Azure AI Services Blog},
    url = {https://techcommunity.microsoft.com/t5/ai-azure-ai-services-blog/comparing-gpt-3-5-amp-gpt-4-a-thought-framework-on-when-to-use/ba-p/4088645}
}

@misc{zhao2023survey,
      title={{A Survey of Large Language Models}}, 
      author={Wayne Xin Zhao and Kun Zhou and Junyi Li and Tianyi Tang and Xiaolei Wang and Yupeng Hou and Yingqian Min and Beichen Zhang and Junjie Zhang and Zican Dong and Yifan Du and Chen Yang and Yushuo Chen and Zhipeng Chen and Jinhao Jiang and Ruiyang Ren and Yifan Li and Xinyu Tang and Zikang Liu and Peiyu Liu and Jian-Yun Nie and Ji-Rong Wen},
      year={2023},
      eprint={2303.18223},
      archivePrefix={arXiv},
      primaryClass={cs.CL}
}

@misc{Google:PaLM:2024,
    author = {{Google}},
    title = {{PaLM Documentation}},
    year = {2024},
    url = {https://ai.google.dev/palm\_docs/palm},
    note = {Accessed: 2024-07-29}
}

@misc{DatabricksBlog2023DollyV2,
    author    = {Mike Conover and Matt Hayes and Ankit Mathur and Jianwei Xie and Jun Wan and Sam Shah and Ali Ghodsi and Patrick Wendell and Matei Zaharia and Reynold Xin},
    title     = {{Free Dolly: Introducing the World's First Truly Open Instruction-Tuned LLM}},
    year      = {2023},
    url       = {https://www.databricks.com/blog/2023/04/12/dolly-first-open-commercially-viable-instruction-tuned-llm},
    urldate   = {2023-06-30}
}

@InProceedings{Pythia:2023, 
author={Stella Biderman and Hailey Schoelkopf and Quentin Anthony and Herbie Bradley and Kyle O'Brien and Eric Hallahan and Mohammad Aflah Khan and Shivanshu Purohit and USVSN Sai Prashanth and Edward Raff and Aviya Skowron and Lintang Sutawika and Oskar van der Wal},
title={{Pythia: A Suite for Analyzing Large Language Models Across Training and Scaling}},
booktitle="The 40th International Conference on Machine
Learning",
year="2023",
address="Honolulu, Hawaii, USA",
}

@misc{anil2023palm,
      title={{PaLM 2 Technical Report}}, 
      author={Rohan Anil and Andrew M. Dai and Orhan Firat and Melvin Johnson and Dmitry Lepikhin and Alexandre Passos and Siamak Shakeri and Emanuel Taropa and Paige Bailey and Zhifeng Chen and Eric Chu and Jonathan H. Clark and Laurent El Shafey and Yanping Huang and Kathy Meier-Hellstern and Gaurav Mishra and Erica Moreira and Mark Omernick and Kevin Robinson and others},
      year={2023},
      eprint={2305.10403},
      archivePrefix={arXiv},
      primaryClass={cs.CL}
}

@misc{PaLM2-Blog,
  author = {Sharan Narang and Aakanksha Chowdhery},
  title = {{Pathways Language Model (PaLM): Scaling to 540 Billion Parameters for Breakthrough Performance}},
  year = {2022},
  month = {April},
  url = {https://blog.research.google/2022/04/pathways-language-model-palm-scaling-to.html},
  urldate = {2024-02-02}
}

@misc{touvron2023llama,
      author={Hugo Touvron and Thibaut Lavril and Gautier Izacard and Xavier Martinet and Marie-Anne Lachaux and Timothée Lacroix and Baptiste Rozière and Naman Goyal and Eric Hambro and Faisal Azhar and Aurelien Rodriguez and Armand Joulin and Edouard Grave and Guillaume Lample},
      title={{LLaMA: Open and Efficient Foundation Language Models}}, 
      year={2023},
      eprint={2302.13971},
      archivePrefix={arXiv},
      primaryClass={cs.CL}
}
%

\end{document}